\documentclass[letterpaper, 10 pt, conference]{ieeeconf}

\IEEEoverridecommandlockouts
\usepackage{amsmath}
\usepackage{amssymb}
\usepackage{microtype}
\usepackage{graphicx}
\usepackage{booktabs}
\usepackage{multirow}
\usepackage{float}
\usepackage{cuted}
\usepackage{capt-of}
\usepackage[hidelinks]{hyperref}

\usepackage[table]{xcolor}
\definecolor{bestcolor}{gray}{.9}
\newcommand{\bestcell}[1]{\cellcolor{bestcolor}{#1}}
\usepackage{makecell}

\makeatletter
\long\def\@makecaption#1#2{%
\ifx\@captype\@IEEEtablestring%
\noindent\parbox[t]{\hsize}{\footnotesize #1.~~\normalfont #2}\par%
\@IEEEtablecaptionsepspace%
\else
\@IEEEfigurecaptionsepspace%
\setbox\@tempboxa\hbox{\footnotesize #1.~~ #2}%
\ifdim \wd\@tempboxa >\hsize%
\setbox\@tempboxa\hbox{\footnotesize #1.~~ }%
\parbox[t]{\hsize}{\footnotesize \noindent\unhbox\@tempboxa#2}%
\else%
\ifcenterfigcaptions \hbox to\hsize{\footnotesize\hfil\box\@tempboxa\hfil}%
\else \hbox to\hsize{\footnotesize\box\@tempboxa\hfil}%
\fi\fi\fi}
\makeatother
\title{\LARGE \bf
Immiscible Diffusion Policy: Preserving Multimodal Robot Actions through Label-Free Noise Assignment
}

\author{%
Xiao Zhang$^{1,*}$, Yuxin Chen$^{1,*}$, Zhixuan Liang$^{2,*}$,
Guojian Zhan$^{1}$, Chenran Li$^{1}$,\\
Chenfeng Xu$^{3}$, Masayoshi Tomizuka$^{1}$, Yiheng Li$^{1}$\\[0.5em]
{\small\itshape
$^{1}$University of California, Berkeley \quad
$^{2}$Princeton University \quad
$^{3}$The University of Texas at Austin}
\thanks{$^{*}$Equal contribution.}%
\thanks{Project page:
\href{https://immiscible-diffusion-policy.github.io}{immiscible-diffusion-policy.github.io}}%
\thanks{Code:
\href{https://github.com/immiscible-diffusion-policy/immiscible-diffusion-policy}{github.com/immiscible-diffusion-policy/immiscible-diffusion-policy}}%
}

\newcommand{\humanoidtasksfigure}{%
\begin{figure}[H]
    \centering
    \includegraphics[width=0.9\columnwidth]{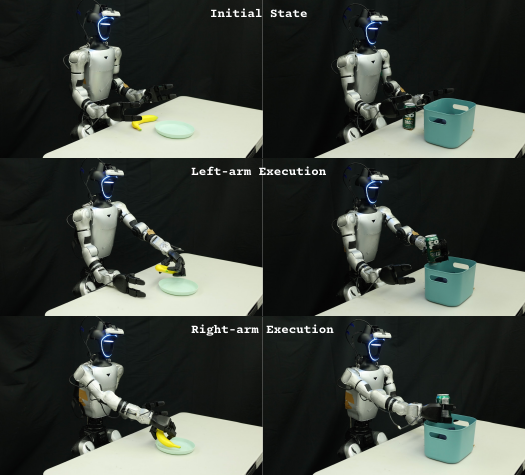}
    \caption{\textbf{Real-world humanoid tasks.} Left: Fruit-to-Plate; right: Can-to-Bin. Rows show the initial state and actions using either arm, corresponding to the two valid modalities in each task.}
    \label{fig:humanoid_tasks}
    \vspace{-0.4cm}\end{figure}%
}

\newcommand{\mainresultstable}{%
\begin{table*}[tb]
\centering
\caption{\textbf{Action-modality preservation and task performance.} Immiscible Diffusion Policy consistently produces more balanced modality distributions while maintaining strong task performance. Mean $\pm$ standard error over $N$ seeds, except TV (mean only). TV is the total variance distance to a balanced distribution ($\downarrow$); modes are ordered by vanilla-policy frequency. Push-T reports coverage; all other tasks report success.}
\label{tab:main_multimodality_results}
\setlength{\tabcolsep}{2pt}

\textbf{(a) Two-modality tasks}\\[4pt]
\resizebox{0.85\textwidth}{!}{%
\scriptsize
\begin{tabular}{lcc|ccccc}
\toprule
\textbf{Task} & \textbf{N} & \textbf{Expert data (\%)} & \textbf{Policy} & \textbf{Dominant (\%)} & \textbf{Non-dominant (\%)} &
\textbf{TV (\%) $\downarrow$} & \textbf{Task metric (\%)} \\
\midrule
\multirow[c]{2}{*}{Push-T} & \multirow[c]{2}{*}{10} & \multirow[c]{2}{*}{$(43.1,56.9)$} & Vanilla
& $97.60 \pm 1.62$ & $2.40 \pm 1.62$
& $47.60$ & $95.16 \pm 0.30$ (Coverage) \\
& & & \bestcell{\textbf{Immiscible}}
& \bestcell{$65.00 \pm 7.64$} & \bestcell{${35.00 \pm 7.64}$}
& \bestcell{${15.00}$} & \bestcell{$95.39 \pm 0.10$ (Coverage)} \\
\multirow[c]{2}{*}{Stack Blocks} & \multirow[c]{2}{*}{3} & \multirow[c]{2}{*}{$(50.0,50.0)$} & Vanilla
& $96.67 \pm 2.72$ & $3.33 \pm 2.72$
& $46.67$ & $96.67 \pm 2.72$ (Success) \\
& & & \bestcell{\textbf{Immiscible}}
& \bestcell{$80.00 \pm 0.00$} & \bestcell{${20.00 \pm 0.00}$}
& \bestcell{${30.00}$} & \bestcell{$96.67 \pm 2.72$ (Success)} \\
\multirow[c]{2}{*}{Place Bread} & \multirow[c]{2}{*}{3} & \multirow[c]{2}{*}{$(51.0,49.0)$} & Vanilla
& $95.56 \pm 3.63$ & $4.44 \pm 3.63$
& $45.56$ & $100.00 \pm 0.00$ (Success) \\
& & & \bestcell{\textbf{Immiscible}}
& \bestcell{$66.67 \pm 8.31$} & \bestcell{${33.33 \pm 8.31}$}
& \bestcell{${16.67}$} & \bestcell{$97.78 \pm 1.81$ (Success)} \\
\midrule
\multirow[c]{2}{*}{\makecell[l]{Humanoid\\Fruit-to-Plate}} & \multirow[c]{2}{*}{1} & \multirow[c]{2}{*}{$(50.0,50.0)$} & Vanilla
& $100$ & $0$
& $50$ & $75$ (Success)\\
& & & \bestcell{\textbf{Immiscible}}
& \bestcell{$50$} & \bestcell{${50}$}
& \bestcell{${0}$} & \bestcell{$60$ (Success)} \\
\multirow[c]{2}{*}{\makecell[l]{Humanoid\\Can-to-Bin}} & \multirow[c]{2}{*}{1} & \multirow[c]{2}{*}{$(50.0,50.0)$} & Vanilla
& $100$ & $0$
& $50$ & $100$ (Success)\\
& & & \bestcell{\textbf{Immiscible}}
& \bestcell{$50$} & \bestcell{${50}$}
& \bestcell{${0}$} & \bestcell{$90$ (Success)} \\
\bottomrule
\end{tabular}}
\par\medskip\textbf{(b) Four-modality tasks}\\[4pt]
\resizebox{\textwidth}{!}{%
\scriptsize
\begin{tabular}{lcc|ccccccc}
\toprule
\textbf{Task} & \textbf{N} & \textbf{Expert data (\%)} & \textbf{Policy}
&  $\boldsymbol{M_{(1)}}$ (\%) & $\boldsymbol{M_{(2)}}$ (\%)
& $\boldsymbol{M_{(3)}}$ (\%) & $\boldsymbol{M_{(4)}}$ (\%)
& \textbf{TV (\%) $\downarrow$} & \textbf{Task metric (\%)} \\
\midrule
\multirow[c]{2}{*}{Place Shoe} & \multirow[c]{2}{*}{3} & \multirow[c]{2}{*}{$(25.7,23.8,24.3,26.2)$} & Vanilla
& $37.80 \pm 14.50$ & $37.80 \pm 15.50$
& $24.40 \pm 1.80$ & $0.00 \pm 0.00$
& $38.33$ & $100.00 \pm 0.00$ (Success) \\
& & & \bestcell{\textbf{Immiscible}}
& \bestcell{$24.92 \pm 1.75$} & \bestcell{$29.68 \pm 4.89$}
& \bestcell{$29.37 \pm 5.53$} & \bestcell{${16.03 \pm 8.16}$}
& \bestcell{${13.57}$} & \bestcell{$93.33 \pm 3.14$ (Success)} \\
\multirow[c]{2}{*}{\makecell[l]{Rotate\\QR Code}} & \multirow[c]{2}{*}{3} & \multirow[c]{2}{*}{$(23.3,24.8,26.7,25.2)$} & Vanilla
& $57.80 \pm 11.90$ & $35.60 \pm 12.70$
& $6.70 \pm 3.10$ & $0.00 \pm 0.00$
& $49.44$ & $86.67 \pm 5.44$ (Success) \\
& & & \bestcell{\textbf{Immiscible}}
& \bestcell{$44.40 \pm 9.60$} & \bestcell{$17.80 \pm 6.50$}
& \bestcell{$26.70 \pm 6.30$} & \bestcell{${11.10 \pm 6.50}$}
& \bestcell{${28.33}$} & \bestcell{$73.33 \pm 3.14$ (Success)} \\
\bottomrule
\end{tabular}}
\vspace{-0.4cm}\end{table*}%
}

\newcommand{\noiseorganizationfigures}{%
\begin{figure}[H]
    \centering
    \includegraphics[width=\columnwidth]{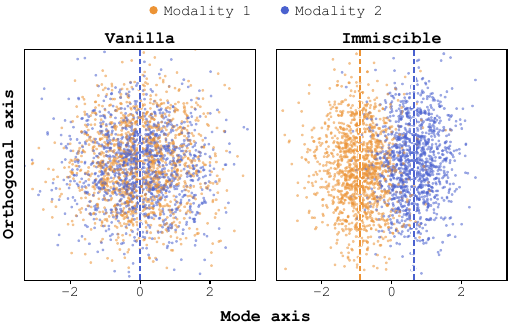}
    \par
    \caption{\textbf{Noise-space organization on Push-T.} Noise samples are projected onto modality-related and orthogonal axes and colored by paired-action modality. The orthogonal axis has no specific meaning and is included only for visualization. Hungarian assignment yields clearer modality-dependent structure than independent pairing.}
    \label{fig:pusht_noise_space}
    \end{figure}
\begin{figure*}[t]
    \centering
    \includegraphics[width=\textwidth]{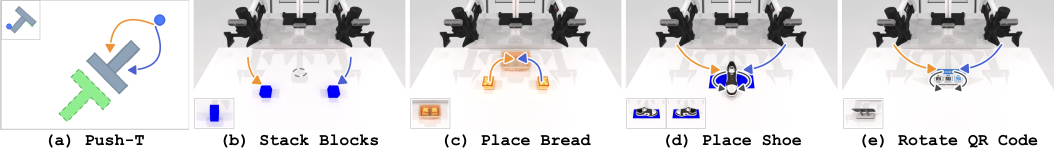}
    \caption{Overview of the five simulated manipulation tasks. From left to right: Push-T, Stack Blocks, Place Bread, Place Shoe, and Rotate QR Code.}
    \label{fig:task_overview}
    \vspace{-0.4cm}\end{figure*}%
}

\newcommand{\modalitysuccesstable}{%
\begingroup
\setlength{\intextsep}{8pt}\begin{table}[H]\centering
    \caption{\textbf{Modality-wise success rate.} Values are mean $\pm$ standard error over 3 seeds; Overall includes all rollouts, and ``-'' denotes no generated rollout. Modality indices follow Tab.~\ref{tab:main_multimodality_results}.}
    \label{tab:recovered_modality_success}
    \setlength{\tabcolsep}{2.5pt}
    \footnotesize
    \begin{tabular}{@{}lcccc@{}}
        \toprule
        & \multicolumn{2}{c}{Place Shoe}
        & \multicolumn{2}{c}{Rotate QR Code} \\
        \cmidrule(lr){2-3}\cmidrule(l){4-5}
        & Vanilla & Immiscible & Vanilla & Immiscible \\
        \midrule
        $M_{(1)}$ & $100.00 \pm 0.00$ & $90.91 \pm 8.67$
                  & $88.46 \pm 6.27$ & $85.00 \pm 7.98$ \\
        $M_{(2)}$ & $100.00 \pm 0.00$ & $100.00 \pm 0.00$
                  & $87.50 \pm 8.27$ & $87.50 \pm 11.69$ \\
        $M_{(3)}$ & $100.00 \pm 0.00$ & $100.00 \pm 0.00$
                  & $66.67 \pm 27.22$ & $58.33 \pm 14.23$ \\
        $M_{(4)}$ & - & $85.71 \pm 13.23$
                  & - & $40.00 \pm 21.91$ \\
        \midrule
        Overall & $100.00 \pm 0.00$ & $93.33 \pm 3.14$
                & $86.67 \pm 5.44$ & $73.33 \pm 3.14$ \\
        \bottomrule
    \end{tabular}
    \vspace{-0.1cm}\end{table}
\endgroup%
}

\newcommand{\activationtimingtable}{%
\begin{table}[H]\centering
    \caption{\textbf{Push-T assignment activation timing.}}
    \label{tab:activation_timing}
    \setlength{\tabcolsep}{5.5pt}
    \footnotesize
    \begin{tabular}{@{}c|cc@{}}
        \toprule
        \shortstack{\textbf{Assignment}\\\textbf{Activation Epoch}}
        & \shortstack{\textbf{Dominant Modality}\\\textbf{(\%) $\downarrow$}}
        & \shortstack{\textbf{Coverage}\\\textbf{(\%) $\uparrow$}} \\
        \midrule
        0 & $65.44 \pm 4.42$ & $90.26 \pm 1.26$ \\
        \textbf{50} & $\boldsymbol{65.00 \pm 7.64}$ & $\boldsymbol{95.39 \pm 0.10}$ \\
        100 & $77.31 \pm 4.68$ & $95.82 \pm 0.19$ \\
        \bottomrule
    \end{tabular}
    \vspace{-0.25cm}\end{table}%
}

\begin{document}

\maketitle
\thispagestyle{empty}
\pagestyle{empty}

\begin{strip}
    \vspace*{-1.46cm}
    \centering
    \begin{minipage}{\textwidth}
        \centering
        \includegraphics[width=\linewidth]{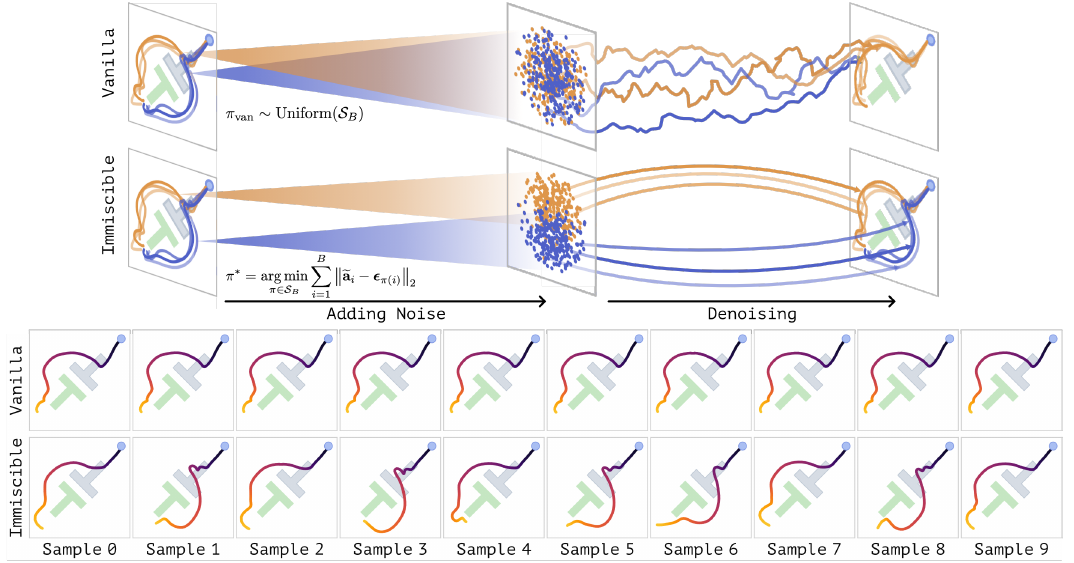}
        \par
        \captionof{figure}{\textbf{Action-noise assignment and action-modality
        preservation in Push-T.} \textbf{Upper:} in low-dimensional action spaces, diffusion paths from different modalities can substantially overlap. Random action-noise pairing in vanilla Diffusion Policy increases path mixing and weakens modality-specific denoising responses. Our batch-wise Hungarian assignment instead introduces modality-dependent organization in the noise space, keeping multiple action modalities reachable without modality labels. \textbf{Lower:} vanilla policy generates trajectories in a single modality, while ours produces a balanced $1{:}1$ distribution. Both policies are trained with the same seed and sampled ten times from a fixed observation. Each column uses the same initial noise.}
        \label{fig:pusht_sample_grid}
        \end{minipage}
\end{strip}

\begin{abstract}
When diffusion policies were first introduced, they were expected to recover multi-modal action distributions. However, we find this expectation does not always hold as diffusion policies often collapse to a single modality even when we guarantee the balance of dataset modalities and exact within-batch symmetry. Our analysis indicates that independent action-noise pairing contributes to this failure by increasing mixing and crossing among diffusion paths (the ``miscibility" in \cite{li2024immiscible}), which can produce averaged denoising responses and suppress modality-specific behavior. It is especially severe in robot planning, as action spaces are dense and low-dimensional, significantly increasing such mixing and crossing. To alleviate this problem, we propose \textit{Immiscible Diffusion Policy}, a label-free training-time add-on to diffusion policy that uses action-noise assignment to preserve relative distinct noise-to-action routes without modifying policy architecture or inference. Across five simulated and two real-world humanoid manipulation tasks spanning state, RGB, and point-cloud observations, our method significantly improves the policy's modality preservation while maintaining strong task performance. It increases the non-dominant-modality proportion by $6.0\times$-$14.6\times$ across three two-modality tasks and recovers demonstrated modalities entirely absent from vanilla policy roll-outs on both four-modality tasks. These results demonstrate that \textit{Immiscible Diffusion Policy} provides a simple yet robust approach to preserving action multi-modality in general robot learning tasks.
\end{abstract}

\section{Introduction}

Diffusion-based policies have achieved strong performance in robotic manipulation by modeling complex action distributions through iterative denoising~\cite{chi2023diffusionpolicy}. Because different initial-noise samples can generate different action trajectories, diffusion policies are heuristically expected to preserve multiple valid action modalities. In practice, however, we find that this expectation does not always hold. For example, in Push-T, the T-shaped object is initialized $180^\circ$ from its target orientation, creating two symmetric action modalities: clockwise and counterclockwise pushing. As shown in the lower rollout comparison of Fig.~\ref{fig:pusht_sample_grid}, a vanilla Diffusion Policy is sampled ten times from this fixed observation, yet all ten generated trajectories follow a single modality. We further repeat this experiment across 10 independent training seeds, enforcing exact symmetry in every batch by pairing each demonstration with its mirrored counterpart. Nevertheless, all 10 vanilla diffusion policies collapse to a single action mode, with the selected mode varying across seeds, as shown in Tab.~\ref{tab:main_multimodality_results} (a). This collapse does more than reduce behavioral diversity; it can eliminate alternative actions that become essential when the initially preferred mode is infeasible at deployment.

Preserving action multimodality is therefore important for robust and safe execution under distribution shifts. Because robotic training data cannot cover every deployment condition, changes in obstacles, workspace boundaries, object configurations, or execution constraints may invalidate one demonstrated action mode while leaving another viable. A policy that has collapsed to the invalid mode may consequently fail despite having observed a feasible alternative during training~\cite{li2024ddiffpg}. This issue also arises in data-rich domains such as autonomous driving, where rare but consequential situations motivate predicting multiple plausible futures for downstream planning~\cite{motionlm}.
Existing approaches often rely on mode-specific supervision, predefined mode structures, or modifications to the policy architecture or inference procedure, making them difficult to apply when action modes are continuous, numerous, or unknown in advance. We therefore seek a label-free method that preserves action modality in existing diffusion policies without requiring the number or semantics of the underlying modes to be specified.

To address this problem, we propose \emph{Immiscible Diffusion Policy}, a label-free training-time add-on for preserving action multimodality in diffusion policies. Inspired by~\cite{li2024immiscible}, we achieve this by explicit data-noise coupling that helps reduce the mixing and crossing of diffusion paths, thereby keeping each action modality reachable from distinct regions of the Gaussian noise space. We use Hungarian matching to couple action chunks with noise samples during training while leaving the policy architecture and inference procedure unchanged. As illustrated in Fig.~\ref{fig:pusht_sample_grid}, the assignment organizes noise into modality-correlated regions that preserve distinct denoising paths. Across all seven evaluated tasks, Immiscible Diffusion Policy consistently improves action-modality preservation while maintaining strong task performance. Our policy increases the non-dominant-modality proportion in the three simulated two-modality tasks by $6.0\times$-$14.6\times$. It further recovers demonstrated modalities entirely absent from vanilla policy rollouts on both four-modality tasks and produces a balanced $1{:}1$ distribution on two real-world humanoid tasks. Our analyses and ablations further verify that action-modality collapse is strongly connected to diffusion-path mixing in low-dimensional, clustered action spaces and show that action-noise assignment preserves the modality-specific denoising responses needed to keep demonstrated action modalities reachable.
Our contributions are summarized as follows:
\begin{enumerate}
    \item We identify and systematically characterize action-modality collapse in diffusion policies, showing that it occurs even with balanced demonstrations and exact within-batch symmetry. Theoretical and empirical analyses connect this failure to denoising averaging induced by diffusion-path mixing and crossing in low-dimensional, clustered action spaces.

    \item We introduce Immiscible Diffusion Policy, a label-free, training-time-only add-on to diffusion policy that preserves action multimodality without requiring additional modality annotations or modifying the policy architecture and inference procedure.

    \item We demonstrate the effectiveness of Immiscible Diffusion Policy across diverse robotic tasks in simulation and the real world, different policy architectures including Diffusion Policy and 3D Diffusion Policy, and different observation representations. Further ablations show it preserves or restores non-dominant modalities that collapse under the vanilla policy, while keeping the dominant modalities' success rates almost intact.

\end{enumerate}

\section{Related Work}

\subsection{Action Multimodality in Robot Learning}

Multiple task-equivalent action sequences often exist for the same observation in robot learning~\cite{florence2021ibc,shafiullah2022bet}. Preserving action's multimodality maintains the behavioral diversity present in demonstrations and can improve robustness by providing alternative strategies under environmental changes or perturbations~\cite{lynch2020playlmp,li2024ddiffpg}.

To keep this multimodality, prior methods often represent diverse behaviors through different policy formulations. Latent-variable approaches, including Play-LMP and ACT, condition actions on learned latent plans but require a dedicated latent representation~\cite{lynch2020playlmp,zhao2023act}. Implicit Behavioral Cloning represents alternative actions as distinct low-energy regions at the cost of iterative inference-time optimization~\cite{florence2021ibc}. Behavior Transformer and VQ-BeT instead capture behavioral modes through explicit action discretizations~\cite{shafiullah2022bet,lee2024vqbet}.

Diffusion Policy models multimodal action distributions by denoising sampled Gaussian noise into action sequences and has demonstrated both short- and long-horizon multimodality~\cite{chi2023diffusionpolicy}. This formulation has been extended to point-cloud and goal-conditioned policies~\cite{ze2024dp3,reuss2023beso}. Efficiency-oriented alternatives include Hybrid Consistency Policy and Consistency Policy, with the latter reporting reduced Push-T multimodality after distillation~\cite{zhao2026hybrid,prasad2024consistency}. IMLE Policy~\cite{rana2025imle} studies multimodal action generation under limited demonstrations using a single-step alternative, while Deep Diffusion Policy Gradient~\cite{li2024ddiffpg} explicitly discovers behavior modes for mode-aware online training. Our work identifies a different failure in the standard Diffusion Policy: demonstrated modalities can be suppressed or lost during training despite balanced multimodal data without explicit modality introduction in the model. Our Immiscible Diffusion Policy addresses this collapse through training-only, label-free action-noise assignment, without modifying the policy architecture or inference procedure, requiring pre-specified modalities, or limiting the number of covered modalities.

\subsection{Data-Noise Assignment in Diffusion Models}

Prior work has explored data-noise assignment to simplify the generative transport learned by diffusion and flow-based models. In flow matching, mini-batch optimal-transport coupling assigns source noise to target data to obtain straighter paths and improve training or sampling efficiency~\cite{tong2024otcfm,pooladian2023multisample}. Immiscible Diffusion similarly introduces structured image-noise assignment to reduce diffusion-path mixing~\cite{li2024immiscible}. Improved Immiscible Diffusion further shows that reducing such miscibility simplifies denoising and accelerates diffusion training, primarily in high-dimensional, sparsely distributed image spaces~\cite{li2026improved}.

Robot planning operates in a substantially lower-dimensional and denser space than image generation. Consequently, action clusters corresponding to distinct strategies can lie close together and overlap across a broad range of noise levels, making diffusion-path mixing and the resulting modality collapse substantially more pronounced. Motivated by this image-noise coupling principle, our method introduces batch-level action-noise assignment to preserve demonstrated action modalities in conditional diffusion policies. Unlike prior coupling methods that primarily target generative efficiency, our objective is action multimodality preservation, while the policy architecture, noise schedule, and test-time sampler remain unchanged.

\begingroup
\section{Diagnosing Action-Modality Collapse}

\subsection{Modality Collapse in Diffusion Policy}

Diffusion models have demonstrated a remarkable ability to capture multimodal distributions, establishing themselves as a dominant paradigm for modern image, audio, and video generation~\cite{ho2020ddpm,kong2021diffwave,ho2022videodiffusion} and motivating their adoption in planning. For this reason, Diffusion Policy is expected to recover multimodality in action distributions, as different initial Gaussian noise samples can intuitively be denoised into diverse action trajectories~\cite{chi2023diffusionpolicy}. Nevertheless, our experiments show that this expectation does not always hold. As shown in the lower part of Fig.~\ref{fig:pusht_sample_grid}, both clockwise and counterclockwise pushing can rotate the T-shaped object to its target orientation. However, even when we guarantee that both action modalities are equally represented in every training batch, each trained policy can concentrate nearly all of its empirical rollouts on a single direction. This collapse occurs across all 10 training seeds. Moreover, different seeds may favor different directions, indicating that the imbalance does not arise from an inherent geometric preference for one modality. These results demonstrate that balanced demonstrations alone do not guarantee that Diffusion Policy will preserve the corresponding action modalities.

\subsection{Why Action-Modality Collapse Happens}

To understand this collapse, we first ask whether it is an unavoidable consequence of the diffusion formulation or a failure of the learned denoiser. Following the analytical denoising formulation of Karras et al.~\cite{karras2022edm}, we construct a reference denoiser directly from expert Push-T action chunks under the same observation. When sampled with the same finite-step reverse process, the analytical denoiser produces action chunks corresponding to both clockwise and counterclockwise modalities at an approximately balanced ratio. In contrast, the trained vanilla policies assign an average of $97.6\%$ of their rollouts to the modality preferred by each seed. Eight of the 10 policies generate only one modality across all 50 evaluation rollouts. Finite-step diffusion sampling can preserve both modalities, whereas the learned vanilla denoiser often does not.

The difference is also apparent in how the generated action depends on the initial noise. Given the same 256 initial noise samples and the same sampler, the nearest-expert assignments of the samples produced by the analytical denoiser cover 174 distinct expert action chunks distributed across both modalities. By contrast, vanilla denoiser outputs cover only 2.6 distinct expert action chunks on average across the 10 seeds, with three seeds covering only one, indicating substantial contraction in action space. Together with the single-modality rollouts in Tab.~\ref{tab:main_multimodality_results}, this shows that diverse initial noise maps to few actions within one modality.

We relate this loss of noise sensitivity to the difficulty of learning modality-specific denoising responses under independent action-noise pairing. In standard diffusion training, each action chunk, a sequence of future actions conditioned on the current observation, is independently paired with Gaussian noise~\cite{ho2020ddpm,chi2023diffusionpolicy}. This random pairing provides no consistent association between sampled noise and its action target. Because robot actions are relatively low-dimensional and action chunks often form compact clusters, different modalities' action representations can be quite close and therefore substantially overlap after noise is added. Similar noisy inputs may therefore correspond to actions from different modalities, causing the MSE-optimal prediction in the overlapping region to approach their conditional average rather than either modality~\cite{florence2021ibc,aithal2024modeinterpolation}.

This conditional averaging does not itself imply collapse: the analytical denoiser can separate the shared response into both modalities as the noise level decreases. Preserving both modalities therefore requires the learned denoiser to accurately capture the transition from this shared region to distinct modality-specific predictions. Independent pairing provides no stable modality-related structure in the noise to facilitate this separation, making the required response difficult to learn from finite data. When this response remains too weak, the finite learned reverse process may fail to separate the two branches before denoising is completed. Small asymmetries introduced by stochastic optimization can consequently bias the learned denoiser toward one branch~\cite{summers2021nondeterminism}. Sampling may propagate this bias throughout the reverse process~\cite{raya2023symmetry}, while closed-loop dynamics can further amplify its effect during execution. As a result, most initial noise samples may eventually produce the same action modality.

This mechanism predicts that a collapsed denoiser should respond weakly to noise variations that distinguish different action modalities. To test this prediction, we vary the noisy input along the direction connecting the two action-modality centroids and measure the corresponding change in the predicted clean action. We refer to this quantity as the modality-response range: a large range indicates that the denoiser distinguishes inputs associated with different modalities, whereas a small range indicates that these differences are suppressed. At the noise level where the two modalities begin to separate, the analytical denoiser has a response range of 4.29, while the trained vanilla denoiser reaches only 0.24, or $5.6\%$ of the analytical response. This flattened response confirms that the vanilla denoiser is largely insensitive to the noisy-input differences required to route samples toward distinct action modalities, explaining why diverse initial noise samples converge to the same modality.

\section{Method}

\subsection{Immiscible Diffusion Policy}

The preceding analysis relates action-modality collapse to the unstructured coupling between action chunks and Gaussian noise. If different action modalities are associated with distinguishable regions of the noise space during training, the denoiser will no longer need to split them from a modality-shared region. Specifically, we note that explicit data-noise coupling helps reduce the mixing and crossing of diffusion paths, thereby keeping each action modality reachable from distinct regions of the Gaussian noise space. This mechanism is illustrated in the upper schematic of Fig.~\ref{fig:pusht_sample_grid}.

Motivated by this observation, we adapt Immiscible Diffusion, originally introduced by Li et al.~\cite{li2024immiscible} to accelerate image diffusion training, to conditional action generation. We refer to the resulting policy-training approach as \emph{Immiscible Diffusion Policy}. Let $\mathbf{a}_i \in \mathbb{R}^{H \times d_a}$ denote a dataset-normalized action chunk of horizon $H$ and action dimension $d_a$, conditioned on observation $\mathbf{o}_i$. For a mini-batch of $B$ action chunks, we independently sample an equally sized noise pool, where each noise sample $\boldsymbol{\epsilon}_j \in \mathbb{R}^{H \times d_a}$ has the same shape as an action chunk:
\begin{equation}
    \boldsymbol{\epsilon}_j
    \sim \mathcal{N}(\mathbf{0},\mathbf{I}_{Hd_a}),
    \qquad j=1,\ldots,B.
\end{equation}
Vanilla diffusion training pairs each $\mathbf{a}_i$ with an independently sampled $\boldsymbol{\epsilon}_i$. Immiscible Diffusion Policy initially uses the same independent pairing during a vanilla-training warm-up. After the warm-up, it treats the standardized action chunks and sampled noise as two sets and activates batch-wise assignment using Hungarian matching. The activation criterion and its sensitivity are described in Sec.~\ref{sec:activation_timing}. Specifically, each action chunk is first vectorized by stacking its temporal and action dimensions and then standardized coordinate-wise using the mini-batch statistics:
\begin{equation}
    \widetilde{\mathbf{a}}_i
    =
    \frac{
    \mathbf{a}_i-\boldsymbol{\mu}_B
    }{
    \boldsymbol{\sigma}_B
    },
\end{equation}
where $\boldsymbol{\mu}_B$ and $\boldsymbol{\sigma}_B$ are the coordinate-wise mean and standard deviation of the vectorized action chunks in the mini-batch. Let $\mathcal{S}_B$ denote the set of permutations of $B$ elements. The assignment is
\begin{equation}
    \pi^{*}
    =
    \underset{\pi\in\mathcal{S}_B}{\arg\min}
    \sum_{i=1}^{B}
    \left\|
    \widetilde{\mathbf{a}}_i
    -
    \boldsymbol{\epsilon}_{\pi(i)}
    \right\|_2 .
\end{equation}
The matched noise is then used in the standard forward diffusion process:
\begin{equation}
    \mathbf{a}_{t_i,i}
    =
    \sqrt{\bar{\alpha}_{t_i}}\,\mathbf{a}_i
    +
    \sqrt{1-\bar{\alpha}_{t_i}}\,
    \boldsymbol{\epsilon}_{\pi^{*}(i)},
\end{equation}
where $\bar{\alpha}_{t_i}$ denotes the cumulative noise schedule at diffusion step $t_i$. Under the noise-prediction parameterization, the policy predicts $\widehat{\boldsymbol{\epsilon}}_i =
\boldsymbol{\epsilon}_{\theta}(\mathbf{a}_{t_i,i},t_i,\mathbf{o}_i)$ and is
trained using
\begin{equation}
\begin{aligned}
    \mathcal{L}_{\mathrm{IDP}}(\theta)
    &=
    \mathbb{E}_{\mathcal{B},\boldsymbol{\epsilon},\mathbf{t}}
    \left[
    \frac{1}{B}\sum_{i=1}^{B}
    \left\|
    \boldsymbol{\epsilon}_{\pi^{*}(i)}
    -
    \widehat{\boldsymbol{\epsilon}}_i
    \right\|_2^2
    \right].
\end{aligned}
\label{eq:idp_objective}
\end{equation}

This training formulation differs from vanilla diffusion training only in how action chunks and noise samples are paired. The noise-assignment algorithm itself follows Li et al.~\cite{li2024immiscible}. Our focus is its application to robot action distributions and its previously unexplored effect on action-modality preservation. Hungarian matching only permutes the Gaussian noise samples already drawn for each mini-batch. It therefore leaves the batch-level noise marginal unchanged while introducing an explicit coupling between actions and noise. The method requires neither additional demonstrations nor additional gradient updates. The policy architecture, loss form, noise schedule, and test-time sampling procedure also remain unchanged. Consequently, Immiscible Diffusion Policy acts as a training-time add-on that can be applied to existing diffusion-policy architectures.
\noiseorganizationfigures\par%
\subsection{How Noise Assignment Preserves Action Multimodality}

Although Immiscible Diffusion Policy does not use action-modality labels, its assignment naturally organizes noise according to the clustered structure of robot actions. Action chunks from the same modality tend to be closer to one another than those from different modalities. By minimizing the batch-level action-noise distance, Hungarian matching correlates nearby action chunks with coherent regions of the Gaussian noise space. The initial noise can therefore carry information about the corresponding action modality rather than being independent of it.

Figure~\ref{fig:pusht_noise_space} visualizes this effect on Push-T. Under vanilla Diffusion Policy, noise samples associated with the two action modalities substantially overlap and mix in the noise space. With Immiscible Diffusion Policy, the noise samples instead exhibit modality-dependent organization, allowing both action modalities to remain reachable from modality-correlated regions of the noise space without using modality labels during training.

This organization is particularly useful in low-dimensional action spaces, where noisy action distributions from different modalities can overlap across a broad range of noise levels. Under independent action-noise pairing, similar noisy inputs may correspond to conflicting action targets, encouraging averaged denoising responses that suppress modality-specific behavior. Assignment instead makes the noise region informative of the corresponding action modality, allowing the denoiser to maintain distinct modality-specific responses throughout the overlapping regime.

Recall that the modality-response range measures the predicted clean-action change under noisy-input variations along the modality direction. At the modality-separation noise level, Immiscible Diffusion Policy achieves a range of 1.93, compared with 0.24 for vanilla Diffusion Policy. It therefore retains approximately $45\%$ of the analytical denoiser's response, whereas vanilla retains only $5.6\%$. Its response near the midpoint between the two modalities is also approximately $8.3$ times stronger than that of vanilla. These improvements are consistent across all 10 training seeds and show that our policy preserves substantially more of the initial-noise variation required to generate distinct action modalities.

Together, these results support the following mechanism: independent pairing mixes the denoising routes associated with different action modalities and makes the learned policy insensitive to initial noise, whereas our assignment organizes the noise space and maintains distinct noise-to-action routes. By reducing cross-modality mixing, Immiscible Diffusion Policy mitigates action-modality collapse without requiring modality labels or changes to the inference procedure.
\endgroup

\section{Experiments}
\label{sec:experiments}

We structure experiments to answer following questions:
\begin{enumerate}
    \item How severe is action-modality collapse in Diffusion Policy despite modality-balanced demonstrations?
    \item Does Immiscible Diffusion Policy preserve demonstrated action multimodality across different tasks?
    \item Can Immiscible Diffusion Policy preserve multimodality while maintaining task success rates?
\end{enumerate}
\mainresultstable\subsection{Experimental Setup}
\textbf{Evaluation Settings.}
We evaluate multimodality preservation and task performance across all tasks. For each training seed $s$, we conduct $R$ rollouts and compute the empirical frequency of each action mode $m$, denoted by $\hat{p}_s(m)$. We use $R=50$ for Push-T and $R\in\{15,20\}$ for the remaining tasks, depending on the baseline defaults. Every rollout is classified according to its \textit{attempted} mode, irrespective of task success. We report the $\hat{p}_s(m)$, policy performance and the total variation distance~\cite{gibbs2002probabilitymetrics} from the balanced distribution $u_K=(1/K,\ldots,1/K)$:
\begingroup
\begin{equation}
D_{\mathrm{TV}}(\hat{p}_s,u_K)
=
\frac{1}{2}\sum_{m=1}^{K}
\left|\hat{p}_s(m)-\frac{1}{K}\right|,
\label{eq:tv_distance}
\end{equation}
\endgroup
where a lower value indicates better multimodality preservation. Task performance is measured by environment coverage for Push-T, following Diffusion Policy~\cite{chi2023diffusionpolicy}, and success rate for all other tasks. Results are averaged over $N$ training seeds.

\humanoidtasksfigure
\textbf{Tasks.}
We evaluate seven tasks: Push-T~\cite{chi2023diffusionpolicy}, four bimanual RoboTwin~2.0 tasks~\cite{chen2025robotwin}, and two real-world humanoid tasks, shown in Figs.~\ref{fig:task_overview} and~\ref{fig:humanoid_tasks}, respectively. Push-T admits two modes that rotate the T-shaped object clockwise or counterclockwise toward its target. Stack Blocks and Place Bread each admit two modes, defined by which arm first picks up the symmetric block or places its bread into the basket. Place Shoe and Rotate QR Code each admit four modes, combining two arm choices with two valid placement or rotation directions. In real-world Fruit-to-Plate and Can-to-Bin, a Unitree G1 humanoid places a banana on a plate and a soda can in a small trash bin, with two possible modalities respectively. Collectively, these tasks span state, RGB, and point-cloud observations in both two- and four-mode settings. Each task is evaluated from a controlled configuration in which all demonstrated modes are feasible and task-equivalent, with no geometric preference for any mode.

\textbf{Demonstration Data.}
We use 206 expert demonstration episodes for each simulated task and 80 tele-operated demonstration episodes for each real-world task. We further perform data augmentation with mirroring to guarantee the demonstrations are well balanced across all valid action modalities as summarized in Tab.~\ref{tab:main_multimodality_results}. For the simulated tasks, we enforce exact in-batch symmetry across modalities during training by pairing one half of each mini-batch with its mirrored counterpart in the other half. Thus, symmetric action modalities are represented equally in every mini-batch, in addition to being balanced across the complete dataset. The demonstrations for each real-world task are likewise divided equally between the two modalities. The vanilla and Immiscible Diffusion Policies are trained using identical demonstration datasets and mini-batches.

\begin{figure*}[t]
    \centering
    \includegraphics[width=0.88\textwidth]{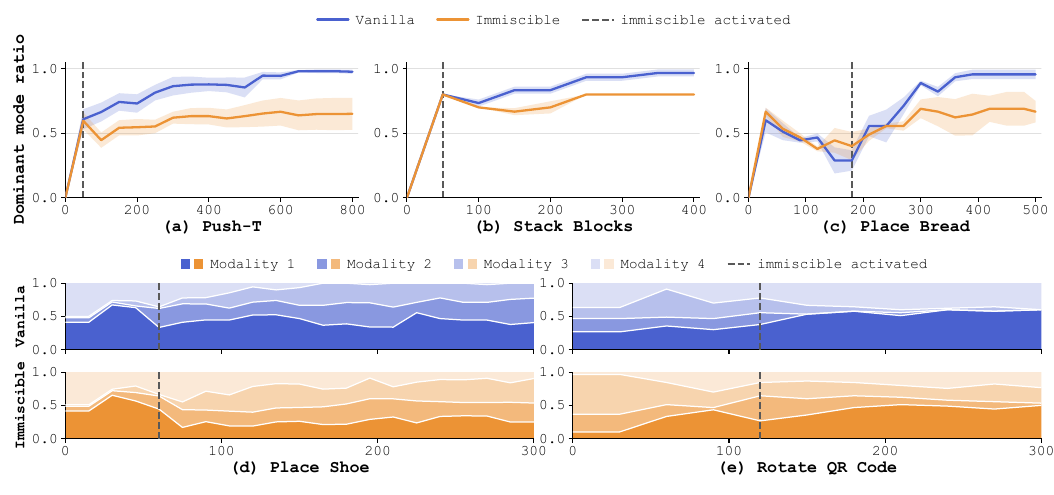}
    \caption{\textbf{Evolution of modality distributions throughout training.} The horizontal axis denotes training epoch. \textbf{Upper:} dominant-modality proportion for three 2-modality tasks, shading denotes standard error; \textbf{Lower:} modality composition for two 4-modality tasks. Results average $N$ seeds from Tab.~\ref{tab:main_multimodality_results}, dashed lines mark assignment activation.}
    \label{fig:modality_training}
    \vspace{-0.4cm}\end{figure*}

\textbf{Baselines and Implementation.}
Our primary baseline is the Diffusion Policy~\cite{chi2023diffusionpolicy}.
The low-dimensional and RGB-observation tasks use the corresponding Diffusion Policy architectures, while the point-cloud tasks use 3D Diffusion Policy~\cite{ze2024dp3}. For each task, the vanilla and Immiscible Diffusion Policies are trained using the same expert demonstrations, policy architecture, observation, action representation, and training settings.
The only methodological difference is the coupling between action chunks and Gaussian noise during training. Immiscible Diffusion Policy activates Hungarian matching using the shared criterion in Sec.~\ref{sec:activation_timing} and recomputes it for each subsequent mini-batch. At inference time, it uses the same test-time sampler, number of denoising steps, and action-execution horizon.

\subsection{Experiment Results}
\textbf{Modality Collapse in Vanilla Diffusion Policy.}
We show the results in Tab.~\ref{tab:main_multimodality_results} and find that despite the well-balanced modality distributions in the expert demonstrations, vanilla Diffusion Policy exhibits pronounced modality collapse across all seven tasks.
The empirical rollout distribution of vanilla Diffusion Policy concentrates almost entirely on a single modality. The dominant-modality proportion reaches $97.60\%$ in Push-T, $96.67\%$ in Stack Blocks, and $95.56\%$ in Place Bread, leaving only $2.40\%$, $3.33\%$, and $4.44\%$ for the non-dominant modality, respectively. Note that the dominant modality is identified for each training seed.

In four-modality tasks, vanilla Diffusion Policy produces the four modalities with frequencies of $37.8\%$, $37.8\%$, $24.4\%$, and $0.0\%$ in Place Shoe, and $57.8\%$, $35.6\%$, $6.7\%$, and $0.0\%$ in Rotate QR Code. The mode collapse is even more severe as one demonstrated modality is entirely absent from the evaluated rollouts in both tasks, while another modality in Rotate QR Code is strongly suppressed to only $6.7\%$.

Finally, on the two real-world humanoid tasks shown in Fig.~\ref{fig:humanoid_tasks}, either arm provides a valid strategy. Nevertheless, vanilla Diffusion Policy maps all evaluated rollouts to a single hand modality on both tasks. Together, these results show that modality collapse occurs across low-dimensional, RGB, and point-cloud observations and in both simulated and real-world environments.

\textbf{Immiscible Diffusion Policy Prevents Modality Collapse.}
As shown in Tab.~\ref{tab:main_multimodality_results}, Immiscible Diffusion Policy consistently redistributes rollout frequency toward modalities suppressed by vanilla Diffusion Policy. In Push-T, the non-dominant-modality proportion increases from $2.40\%$ to $35.00\%$. It increases from $3.33\%$ to $20.00\%$ in Stack Blocks and from $4.44\%$ to $33.33\%$ in Place Bread. Across these three simulated two-modality tasks, our policy increases the average non-dominant-modality proportion by $26.05$ percentage points, corresponding to an $8.68\times$ increase over the vanilla policy. In both real-world humanoid tasks, our policy changes the rollout distribution from a single modality under vanilla Diffusion Policy to a balanced $1{:}1$ split between the left- and right-hand modalities. Our policy therefore transforms the nearly deterministic modality selection of vanilla Diffusion Policy into a substantially more balanced multimodal distribution across both simulated and real-world tasks.

The recovery is also evident in the four-modality tasks. In Place Shoe, Immiscible Diffusion Policy changes the generated distribution from $37.8\%$, $37.8\%$, $24.4\%$, and $0.0\%$ to $24.9\%$, $29.7\%$, $29.4\%$, and $16.0\%$, moving the rollout distribution substantially closer to the balanced expert distribution. In Rotate QR Code, the generated distribution changes from $57.8\%$, $35.6\%$, $6.7\%$, and $0.0\%$ to $44.4\%$, $17.8\%$, $26.7\%$, and $11.1\%$. All four demonstrated modalities are retained by our policy and become substantially more balanced. Most importantly, our policy recovers the modality that is entirely absent from the vanilla rollouts in both four-modality tasks, which significantly boosts the model's modality coverage and robustness.

Figure~\ref{fig:modality_training} reports the evolution of modality usage over training for all five simulated tasks. As training progresses, vanilla Diffusion Policy increasingly concentrates its rollouts on a single modality or a subset of modalities. In contrast, Immiscible Diffusion Policy maintains a more balanced and stable modality distribution, preserving multimodality throughout training.

\textbf{Task Performance.}
As shown in Tab.~\ref{tab:main_multimodality_results}, Immiscible Diffusion Policy maintains approximately the same task performance as vanilla Diffusion Policy on the simulated two-modality tasks. Push-T coverage changes only from $95.16\%$ to $95.39\%$, Stack Blocks success remains unchanged at $96.67\%$, and Place Bread success changes only from $100.00\%$ to $97.78\%$. Our policy substantially improves multimodality preservation on the two-modality tasks without meaningfully affecting task performance. On the real-world tasks, Immiscible Diffusion Policy balances the two hand modalities at $1{:}1$, while the success rate changes from $75\%$ to $60\%$ on Fruit-to-Plate and from $100\%$ to $90\%$ on Can-to-Bin. Immiscible Diffusion Policy also maintains relatively strong task performance on the four-modality tasks, with slight success rate drops mainly caused by the restored modalities (details in Sec.~\ref{sec:analysis_ablation}). In Place Shoe, it achieves a success rate of $93.33\%$, compared with $100.00\%$ for vanilla Diffusion Policy. In Rotate QR Code, it achieves $73.33\%$, compared with $86.67\%$ for vanilla Diffusion Policy. Despite these differences, our policy preserves all four demonstrated action modalities in both tasks and recovers the modalities completely lost by vanilla Diffusion Policy.

\modalitysuccesstable\suppressfloats[t]\subsection{Analysis and Ablation Studies}
\label{sec:analysis_ablation}
\label{sec:activation_timing}
We analyze two questions arising from the main results: what accounts for the task-performance differences on the four-modality tasks, and how sensitive our policy is to assignment activation time. We study them through modality-conditioned success rates and an activation-time ablation on Push-T, respectively.

\textbf{Modality-wise Task Performance}
Tab.~\ref{tab:recovered_modality_success} reports the success rate conditioned on the action modality attempted by each rollout. Vanilla Diffusion Policy generates no rollouts from $M_{(4)}$ in either task, whereas Immiscible Diffusion Policy recovers this missing modality. At the evaluated training checkpoint, the recovered $M_{(4)}$ achieves a conditional success rate of $85.71\%$ in Place Shoe and only $40.00\%$ in Rotate QR Code, indicating that the newly recovered behaviors are still being learned and remain less reliable. On the other three modalities, which are generated by both methods, our policy achieves success rates largely comparable to those of vanilla Diffusion Policy. These results suggest that the lower overall success primarily reflects the additional challenge of retaining and continuing to learn a valid modality that vanilla Diffusion Policy has already discarded, rather than a general degradation in policy execution.

\activationtimingtable
\textbf{Assignment Activation Timing.}
An important practical choice for Immiscible Diffusion Policy is when to transition from vanilla training to assignment-based training. Although assignment can be applied from initialization, modality concentration develops progressively during vanilla training, making it unclear whether assignment is needed throughout the entire optimization process. Conversely, delaying activation for too long may allow demonstrated modalities to become fully suppressed or disappear, shifting the role of assignment from preserving active modalities to recovering ones already lost from the policy's rollout distribution. We therefore evaluate how the activation time affects the final task performance and modality distribution.

As shown in Tab~\ref{tab:activation_timing}, compared with Tab.~\ref{tab:main_multimodality_results} setting of epoch 50, activation at epoch 0 achieves similar modality balance but reduces coverage by $5.13$ percentage points under the same training budget, indicating that a short vanilla warm-up can improve task acquisition without sacrificing modality preservation. Delaying activation to epoch 100 maintains coverage but increases the dominant-modality proportion by $12.31$ percentage points. These results motivate activating assignment after meaningful task competence emerges but before severe modality collapse develops under vanilla training.

We empirically operationalize this principle using a shared criterion across tasks: assignment is activated at the first evaluated checkpoint where the task metric reaches at least $60\%$, provided that every demonstrated modality retains a nonzero empirical rollout frequency. Because the task metrics are normalized to $[0,1]$, this shared, deliberately non-saturated threshold requires substantial progress beyond the initial low-performance regime while leaving sufficient training for assignment. Selecting the first qualifying checkpoint limits further vanilla modality concentration, while the modality-presence condition ensures that assignment begins before modality support is lost. The shared criterion also avoids tuning activation epochs per task.

\section{Conclusion}
\label{sec:conclusion}

In this work, we identify the action-modality collapse as an important limitation of diffusion policy for robotic action generation. We show diffusion policies usually concentrate on a single (or a portion of) action modality even when the demonstrations are balanced and symmetry is enforced within every training batch. Our analysis connects this behavior to independent action-noise pairing, which increases the mixing and crossing of diffusion paths and weakens the denoiser's sensitivity to the modality difference in low-dimensional, clustered action spaces. To address this issue, we introduce Immiscible Diffusion Policy, a label-free training-time add-on that couples action chunks with noise samples without changing the policy architecture or inference procedure. Across simulation and real-world tasks, our method consistently improves action-modality preservation while maintaining strong task performance, preserving suppressed behaviors and recovering valid modalities absent from vanilla policy rollouts. These results demonstrate that explicit data-noise coupling provides a simple and broadly applicable approach to preserving action multimodality in general diffusion-based robot planning.

\textbf{Limitations:} Although our method largely preserves the performance of existing modalities, those recovered modalities might suffer from low success rates during their restoration. Future work will investigate how to achieve equally high-quality planning across all preserved modalities.


\begin{thebibliography}{99}

\bibitem{li2024immiscible}
Y.~Li, H.~Jiang, A.~Kodaira, M.~Tomizuka, K.~Keutzer, and C.~Xu,
``Immiscible diffusion: Accelerating diffusion training with noise assignment,''
in \emph{Advances in Neural Information Processing Systems},
vol.~37, 2024, pp.~90198--90225.

\bibitem{chi2023diffusionpolicy}
C.~Chi \emph{et al.},
``Diffusion Policy: Visuomotor policy learning via action diffusion,''
in \emph{Proc. Robotics: Science and Systems (RSS)},
2023.

\bibitem{li2024ddiffpg}
Z.~Li, R.~Krohn, T.~Chen, A.~Ajay, P.~Agrawal, and G.~Chalvatzaki,
``Learning multimodal behaviors from scratch with diffusion policy gradient,''
in \emph{Advances in Neural Information Processing Systems},
vol.~37, 2024, pp.~38456--38479.

\bibitem{motionlm}
A.~Seff \emph{et al.},
``MotionLM: Multi-agent motion forecasting as language modeling,''
in \emph{Proc. IEEE/CVF Int. Conf. Comput. Vis. (ICCV)},
2023, pp.~8579--8590.

\bibitem{florence2021ibc}
P.~Florence \emph{et al.},
``Implicit behavioral cloning,''
in \emph{Proc. 5th Conf. Robot Learn. (CoRL)},
ser. Proceedings of Machine Learning Research, vol.~164,
2022, pp.~158--168.

\bibitem{shafiullah2022bet}
N.~M. Shafiullah, Z.~Cui, A.~Altanzaya, and L.~Pinto,
``Behavior transformers: Cloning $k$ modes with one stone,''
in \emph{Advances in Neural Information Processing Systems},
vol.~35, 2022, pp.~22955--22968.

\bibitem{lynch2020playlmp}
C.~Lynch \emph{et al.},
``Learning latent plans from play,''
in \emph{Proc. Conf. Robot Learn. (CoRL)},
ser. Proceedings of Machine Learning Research, vol.~100,
2020, pp.~1113--1132.

\bibitem{zhao2023act}
T.~Z. Zhao, V.~Kumar, S.~Levine, and C.~Finn,
``Learning fine-grained bimanual manipulation with low-cost hardware,''
in \emph{Proc. Robotics: Science and Systems (RSS)},
2023.

\bibitem{lee2024vqbet}
S.~Lee, Y.~Wang, H.~Etukuru, H.~J. Kim, N.~M.~M. Shafiullah, and L.~Pinto,
``Behavior generation with latent actions,''
in \emph{Proc. 41st Int. Conf. Mach. Learn. (ICML)},
ser. Proceedings of Machine Learning Research, vol.~235,
2024, pp.~26991--27008.

\bibitem{ze2024dp3}
Y.~Ze, G.~Zhang, K.~Zhang, C.~Hu, M.~Wang, and H.~Xu,
``3D diffusion policy: Generalizable visuomotor policy learning via simple
3D representations,''
in \emph{Proc. Robotics: Science and Systems (RSS)},
2024.

\bibitem{reuss2023beso}
M.~Reuss, M.~Li, X.~Jia, and R.~Lioutikov,
``Goal-conditioned imitation learning using score-based diffusion policies,''
in \emph{Proc. Robotics: Science and Systems (RSS)},
2023.

\bibitem{zhao2026hybrid}
Q.~Zhao \emph{et al.},
``Hybrid consistency policy: Decoupling multi-modal diversity and real-time
efficiency in robotic manipulation,''
\emph{IEEE Robot. Autom. Lett.},
vol.~11, no.~7, pp.~8825--8832, 2026.

\bibitem{prasad2024consistency}
A.~Prasad, K.~Lin, J.~Wu, L.~Zhou, and J.~Bohg,
``Consistency policy: Accelerated visuomotor policies via consistency distillation,''
in \emph{Proc. Robotics: Science and Systems (RSS)},
2024.

\bibitem{rana2025imle}
K.~Rana, R.~Lee, D.~Pershouse, and N.~S{\"u}nderhauf,
``IMLE policy: Fast and sample efficient visuomotor policy learning via
implicit maximum likelihood estimation,''
in \emph{Proc. Robotics: Science and Systems (RSS)},
2025.

\bibitem{tong2024otcfm}
A.~Tong \emph{et al.},
``Improving and generalizing flow-based generative models with minibatch
optimal transport,''
\emph{Trans. Mach. Learn. Res.}, 2024.

\bibitem{pooladian2023multisample}
A.-A.~Pooladian, H.~Ben-Hamu, C.~Domingo-Enrich, B.~Amos, Y.~Lipman, and
R.~T.~Q. Chen,
``Multisample flow matching: Straightening flows with minibatch couplings,''
in \emph{Proc. 40th Int. Conf. Mach. Learn. (ICML)},
ser. Proceedings of Machine Learning Research, vol.~202,
2023, pp.~28100--28127.

\bibitem{li2026improved}
Y.~Li, F.~Liang, D.~Kondratyuk, M.~Tomizuka, K.~Keutzer, and C.~Xu,
``Improved immiscible diffusion: Accelerate diffusion training by reducing
its miscibility,''
in \emph{Proc. Eur. Conf. Comput. Vis. (ECCV)}, 2026.

\bibitem{ho2020ddpm}
J.~Ho, A.~Jain, and P.~Abbeel,
``Denoising diffusion probabilistic models,''
in \emph{Advances in Neural Information Processing Systems},
vol.~33, 2020, pp.~6840--6851.

\bibitem{kong2021diffwave}
Z.~Kong, W.~Ping, J.~Huang, K.~Zhao, and B.~Catanzaro,
``DiffWave: A versatile diffusion model for audio synthesis,''
in \emph{Proc. Int. Conf. Learn. Represent. (ICLR)},
2021.

\bibitem{ho2022videodiffusion}
J.~Ho, T.~Salimans, A.~Gritsenko, W.~Chan, M.~Norouzi, and D.~J. Fleet,
``Video diffusion models,''
in \emph{Advances in Neural Information Processing Systems},
vol.~35, 2022, pp.~8633--8646.

\bibitem{karras2022edm}
T.~Karras, M.~Aittala, T.~Aila, and S.~Laine,
``Elucidating the design space of diffusion-based generative models,''
in \emph{Advances in Neural Information Processing Systems},
vol.~35, 2022, pp.~26565--26577.

\bibitem{aithal2024modeinterpolation}
S.~K. Aithal, P.~Maini, Z.~C. Lipton, and J.~Z. Kolter,
``Understanding hallucinations in diffusion models through mode interpolation,''
in \emph{Advances in Neural Information Processing Systems},
vol.~37, 2024, pp.~134614--134644.

\bibitem{summers2021nondeterminism}
C.~Summers and M.~J. Dinneen,
``Nondeterminism and instability in neural network optimization,''
in \emph{Proc. 38th Int. Conf. Mach. Learn. (ICML)},
ser. Proceedings of Machine Learning Research, vol.~139,
2021, pp.~9913--9922.

\bibitem{raya2023symmetry}
G.~Raya and L.~Ambrogioni,
``Spontaneous symmetry breaking in generative diffusion models,''
in \emph{Advances in Neural Information Processing Systems},
vol.~36, 2023, pp.~66377--66389.

\bibitem{gibbs2002probabilitymetrics}
A.~L. Gibbs and F.~E. Su,
``On choosing and bounding probability metrics,''
\emph{Int. Stat. Rev.},
vol.~70, no.~3, pp.~419--435, 2002.

\bibitem{chen2025robotwin}
T.~Chen \emph{et al.},
``RoboTwin 2.0: A scalable data generator and benchmark with strong domain
randomization for robust bimanual robotic manipulation,''
2025, arXiv:2506.18088.

\end{thebibliography}
\end{document}